\documentclass[twoside]{article}
\usepackage{isolatin1}   % necessary for accents in Thomas' name!
\usepackage{fleqn,espcrc2}
\usepackage[dvips]{graphicx}

\usepackage{graphicx}
\usepackage[figuresright]{rotating}

\newcommand{\AmS}{{\protect\the\textfont2
  A\kern-.1667em\lower.5ex\hbox{M}\kern-.125emS}}

\title{Evolutionary Computing}

\author{A.E. Eiben
\address{Free University Amsterdam -- The Netherlands\\
http://www.cs.vu.nl/$_{\tilde{~}}$gusz/} and 
M. Schoenauer
\address{INRIA Rocquencourt -- France\\
http://www-rocq.inria.fr/fractales/Staff/Schoenauer}
}
       
\begin{document}

\begin{abstract}
Evolutionary computing (EC) is an exciting development in Computer
Science. It amounts to building, applying and studying algorithms
based on the Darwinian principles of natural selection. In this paper
we briefly introduce the main concepts behind evolutionary
computing. We present the main components all evolutionary algorithms
(EA), sketch the differences between different types of EAs and survey
 application areas ranging from optimization,
modeling and simulation to entertainment. 
\vspace{1pc}
\end{abstract}

\maketitle

\section{Introduction}
%%%%%%%%%%%%%%%%%%%%%%
 Surprisingly enough, the idea to apply Darwinian principles to
automated problem solving originates from the fifties, long before the
breakthrough of computers \cite{Fogel:Fossile:98}. During the sixties three
different implementations of this idea have been developed at three
different places. In the USA  Fogel introduced evolutionary
programming \cite{Fogel-pere,DFogel}, while Holland called his
method a genetic algorithm \cite{Holland,Goldberg89}. In Germany
Rechenberg and Schwefel invented evolution strategies
\cite{Rechenberg,Schwefel}. For about 15 years these areas developed
separately; it is 
since the early nineties that they are envisioned as different
representatives (``dialects'') of one technology, called evolutionary
computing. It was also in the early nineties that a fourth stream
following the general ideas has emerged -- genetic programming
\cite{Koza,Banzhaf:book}. The contemporary terminology denotes
the whole field by evolutionary computing and considers evolutionary
programming, evolution strategies, genetic  
algorithms, and genetic programming as sub-areas.

\section{What is an evolutionary algorithm?}
%%%%%%%%%%%%%%%%%%%%%%%%%%%%%%%%%%%%%%%%%%%%
\label{EA}

The common underlying idea behind all these techniques is the same:
given a population of individuals, the environmental pressure causes
natural selection (survival of the fittest) and hereby the fitness of
the population is growing. It is easy to see such a process as
optimization. Given an objective function to be maximized we can
randomly create a set of candidate solutions and use the objective
function as an abstract fitness measure (the higher the better). Based
on this fitness, some of the better candidates are chosen to seed the
next generation by applying recombination and/or
mutation. Recombination is applied to two selected candidates, the
so-called parents, and results one or two new candidates, the
children. Mutation is applied to one candidate and results in one new
candidate. Applying recombination and mutation leads to a set of new
candidates, the offspring. Based on their fitness these offspring
compete with the old candidates for a place in the next generation. 
This process can
be iterated until a solution is found or a previously set time limit
is reached. Let us note that many components of such an evolutionary
process are stochastic. According to Darwin, the emergence of new
species, adapted to their environment, is a consequence of the
interaction between the survival of the fittest mechanism and undirected
variations. Variation operators must be stochastic, the choice on
which pieces of information will be exchanged during recombination, as
well as the changes in a candidate solution during mutation, are random.
On the other hand, selection operators can be either deterministic, or
stochastic. In the latter case fitter individuals have 
a higher chance to be selected than less fit ones, but typically even
the weak individuals have a chance to become a parent or to
survive. The general scheme of an evolutionary algorithm can
be given as follows. \\

\noindent
{\tt
{\bf Initialize} population with random \\
\indent{\bf individuals} (candidate solutions)\\
{\bf Evaluate} (compute fitness of) all individuals\\
WHILE not stop DO \\      
\indent{\bf Select} genitors from parent population\\
\indent Create offspring using \\
\indent \indent {\bf variation} operators on genitors\\
\indent {\bf Evaluate} newborn offspring\\
\indent {\bf Replace} some parents by some offspring\\
OD
}\\

Let us note that this scheme falls in the category of
generate-and-test, also known as trial-and-error, algorithms. The
fitness function represents a heuristic estimation of solution quality
and the search process is driven by the variation operators
(recombination and mutation creating new candidate solutions) and the
selection operators. Evolutionary algorithms (EA) are distinguished
within in the family of generate-and-test methods by being population
based, i.e. process a whole set of candidate 
solutions and by the use of recombination to mix information of two candidate
solutions. 

The aforementioned ``dialects'' of evolutionary computing follow the
above general outlines and differ only in technical details. 

\section{Critical issues}
%%%%%%%%%%%%%%%%%%%%%%%%%
\label{issues}
There are some issues that one should keep in mind when designing and
running an evolutionary algorithm. These considerations concern all of
the ``dialects'', 
and will be discussed here in general, without a specific type of evolutionary
algorithm in mind.

One crucial issue when running an EA is to try to preserve the 
{\em genetic diversity} of the population as long as possible. 
Opposite to many other optimization methods, EAs use a whole
population of individuals -- and this is one of the reasons for their
power. However, if that populations starts to concentrate in a very
narrow region of the search space, all advantages of handling many
different individuals vanish, while the burden of computing their
fitnesses remains. This phenomenon is known as premature convergence.
There are two main directions to prevent this: a priori ensuring
creation of new material, for instance by using a high level of
mutation (see section \ref{mutation}); or a posteriori manipulating
the fitnesses of all individuals to create a bias against being
similar, or close to, existing candidates. A well-known technique is
the so-called  niching mechanism. 

{\em Exploration and exploitation} are two terms often used in EC. Although
crisp definitions are lacking \cite{Eiben:ES98} there has been a lot
of discussion 
about them. The dilemma within an optimization procedure is whether to
search around the best-so-far solutions (as their neighborhood
hopefully contains even better points) or explore some totally
different regions of the search space 
(as the best-so-far solutions might only be local optima). An EA must
be set up in such a way that it solves this dilemma without 
a priori knowledge of the kind of landscape it will have to explore.
The exploitation phase can sometimes be ``delegated'' to some local
optimization procedure, whether called as a mutation operator, or
systematically applied to all newborn individuals, moving them to the
nearest local optimum. In the latter case,
the resulting hybrid algorithm is called a memetic algorithm. 

In general, there are two driving forces behind an EA: selection and variation.
The first one represents a push toward quality and is reducing the genetic
diversity of the population. The second one, implemented by
recombination and mutation operators, represents a push toward
novelty and is increasing genetic diversity. To have an EA work
properly, an appropriate balance between these two forces has to be
maintained. At the moment, however, there is not much theory
supporting practical EA design.  

\section{ Components of evolutionary algorithms}
%%%%%%%%%%%%%%%%%%%%%%%%%%%%%%%%%%%%%%%%%%%%%%%%%%%%
%This section will discuss the components of an EA sketched in
%section\ref{EA}, trying to underline which are responsible for making
%EAs successful.

\subsection{Representation}
%--------------------------
\label{representation}
Solving a given problem with an EA starts with specifying a
representation of the candidate solutions. Such candidate solutions
are seen as phenotypes that can have very complex structures. Applying
variation operators directly to these structures might not be
possible, or easy. Therefore these {\em phenotypes} are represented by
corresponding {\em genotypes}. The standard EC machinery
consists of many off-the-shelf variation operators acting on a
specific genotype space, for instance bit-strings, real-valued
vectors, permutations of integers, or trees. Designing an
EA thus often amounts to choosing one of the standard representations
with the corresponding variation operators in mind. 
However, one strength of EAs is their ability to 
tackle any search space provided that 
initialization and variation operators are available. 
Choosing a
standard option is, therefore, not necessary.

\subsection{Fitness or evaluation function}
%--------------------------------------------------------
Fitness-based selection is the force that represents the drive toward
quality improvements in an EA. Designing the fitness function (or
evaluation function) is therefore crucial. 

The first important feature about fitness computation is that it
represents  99\% of the total computational cost
of evolution in most real-world problems. 
Second, the fitness function very often is the only information about
the problem in the algorithm:  Any available
and usable knowledge about the problem domain
should be used.

\subsection{Representation dependent}
%----------------------------------------------
\label{rep-dep}

\subsubsection{Initialization} 
\label{init}
%...........................
The initial population is usually created by some
random sampling of the search space, generally performed as uniformly
as possible. However, in some cases, uniform sampling might not be
well-defined, e.g. on parse-tree spaces, or on unbounded intervals for
floating-point numbers. 

A common practice also is to {\em
  inoculate} some known good solutions into the initial
population. But beware that no bias is better than a wrong bias!

%In general, the issue of initialization is
%poorly documented and its role and effect on algorithm performance is
%poorly understood.

\subsubsection{Crossover}
%.......................
\label{crossover}
Crossover operators take two (or more) parents and generate 
offspring by exchange of information between the parents.
The underlying idea to explain crossover performance is that 
the good fitness of the parents is somehow due to precise parts
of their genetic material (termed {\em building blocks}), and the
recombining those building blocks will result in an increase in
fitness.

Nevertheless, there are numerous other ways to perform crossover.
For instance, crossing over
two vectors of floating-points values 
can be done by linear combination (with uniformly
drawn weights) of the parents values.
The  idea is that information pertaining to the problem at hand
should be somehow exchanged.

Note that the effect of crossover varies from exploration
when the population is highly diversified to exploitation when it
starts to collapse into a small region of the search space.

\subsubsection{Mutation}
%.......................
\label{mutation}
Mutation operators are stochastic transformations of an individual.
The usual compromise between  exploration and exploitation must be
maintained: large mutations are necessary from theoretical
reasons (it ensures the {\em ergodicity} of the underlying stochastic
process), that translate practically
(it is the only way to reintroduce genetic diversity in the end of
evolution); but of course too much too large mutation transform the
algorithm into a random walk -- so most mutations should generate
offspring close to their parents.
There is no standard general mutation, but general trends are 
to modify the value of a component of the genotype with a
small probability (e.g. flip one bit of a bitstring, or, in case of
real-valued components, add zero-mean Gaussian noise with carefully tuned
standard deviation).

\subsubsection{The historical debate}
%.......................
\label{debate}
There has long been a strong debate about the usefulness of crossover.
The GA community considers crossover to be the essential
variation operator \cite{Holland,Goldberg89}, 
%recombining useful
%{\em building blocks} that will be gradually assemble into a good
%solution of the problem at hand, 
while mutation is only a background
necessity. On the other hand, the historical ES
\cite{Rechenberg,Schwefel} and EP \cite{Fogel-pere} researchers did
not use any crossover at all, and 
even claimed later that it could be harmful \cite{Fogel-Stayton}.

The general agreement nowadays is that the answer is  problem-dependent: 
If there exists a ``semantically meaningful'' crossover for
the problem at hand, it is probably a good idea to use it. But
otherwise mutation alone might be sufficient to find good solutions --
and the resulting algorithm can still be called an Evolutionary
Algorithm. 

\subsection{Representation-independent}
%-----------------------------------------------

\subsubsection{Artificial Darwinism}
%........................................
\label{darwinism}
Darwin's theory states that {\em the fittest individuals reproduce and
  survive}. The {\em evolution engine}, i.e. the two steps of {\bf
  selection} (of some parents to become 
genitors) and {\bf replacement} (of some parents by newborn offspring) are
the artificial implementations of these two selective processes. 
They differ in an essential way:
during selection step, the same parent can be selected many times; during
replacement step, each individual (among parents and offspring) either is
selected, or disappears for ever.

Proportional selection (aka
{\em roulette-wheel}) has long been the most popular selection
operator: each parent has a probability to be 
selected that is proportional to its fitness. However, the difficulty
is to scale the fitness to tune the selection pressure:
even the greatest care will not prevent some super-individual to take
over the population in a very short time. Hence the most widely used
today is tournament selection: to select one individual, $T$
individuals are uniformly chosen, and the best of these $T$ is
returned. Of course, both roulette-wheel and tournament repeatedly act
on the 
same current population, to allow for multiple selection of the very
best individuals.

There are two broad categories of replacement methods: either
the parents and the offspring ``fight'' for survival, or only some
offspring are allowed to survive. Denoting by $\mu$ (resp. $\lambda$)
the number of parents (resp. offspring) as in ES history (section
\ref{ES}), the former strategy is called $(\mu + \lambda)$ and the
latter $(\mu , \lambda)$.
When $\mu=\lambda$, the comma strategy is also known as  {\em generational
  replacement}: all offspring simply replace all parents.
When $\lambda=1$, the (plus!) strategy is then termed
{\em steady-state} and amounts to choosing one parent to be replaced. 

An important point about the evolution engine is the monotonicity of
the best fitness 
along evolution: for instance, ES plus strategies are {\em elitist},
i.e. ensure that the best fitness can only increase from one generation
to another, while the comma strategies are not elitist -- though
elitism can be a posteriori added by retaining the best parent when a
decrease of fitness is foreseen.

\subsubsection{Termination criterion}
%...............................
\label{stopping}
There has been very few theoretical studies about when to stop an
Evolutionary Algorithm. 
The usual stopping criterion is a fixed amount of computing time (or,
almost equivalently, of fitness computations). A slightly more subtle
criterion is to stop when a user-defined amount of time has passed
without improvement of the best fitness in the population.

\subsection{Setting the parameters}
%---------------------------
\label{parameters}

EAs typically have a large number of parameters (e.g.
population size, frequency of 
recombination, mutation step-size,  selective pressure,
\ldots). 
%Despite of claims from the dawn of EC stating that evolution
%is very robust,  it is today widely acknowledged that the parameter
%setting can have a great influence of algorithm performance. 
The main
problem in this respect is that even the individual effect of one
parameter is often unpredictable, let alone the combined influence of
all parameters. Most authors rely on
intensive trials (dozens of independent runs for each possible
parameter setting) to calibrate their algorithms -- an option that is
clearly very time consuming. Another possibility is to use
long-existing statistical techniques like ANOVA.
A specific evolutionary  trend is to let
 the EA calibrate itself to a given problem, while solving
that problem (see section \ref{ES}).  

\subsection{Result analysis}
%...............................
\label{result}
As with any randomized algorithm, the results of a single run of an EA
are meaningless.
%, and only statistical analyses should be provided,
%especially when comparing two different kinds of EAs, or when studying
%the robustness of some particular setting.
A typical experimental analysis will run say 
over more than 15 independent runs (everything equal except the
initial population), and present averages, standard deviations, and
T-test in case of comparative experiments.
 
However, one should distinguish {\em design problems}, where the
goal is to find at least one very good solution once,
from {\em day-to-day optimization} (e.g. control, scheduling,\ldots),
where the  
goal is to consistently find a good solution for different inputs.
In the design context, a high standard deviation is desirable provided
the average result is not too bad. In the optimization context, 
 a good average and a small standard
deviation are mandatory.

\section{Historical dialects}
%%%%%%%%%%%%%%%%%%%%%%%%%%%%
As already quoted, EC arose from independent sources. Of course,
each dialect exhibits a large variety in itself; the short
descriptions here are necessarily restricted to one or two main
variants. 

\subsection{Genetic Algorithms}
%----------------------------------
\label{GA}
The standard GA \cite{Holland,Goldberg89}
can be seen as the combination of bit-string
representation, with bit-exchange crossover (applied with given
probability $p_c$) and bit-flip mutation (applied to every bit with
probability $p_m$),
roulette-wheel selection plus generational replacement (though
steady-state replacement can also be used).  

Note that other versions of EAs using the same evolution engine with
different genotypes (and hence variation operators) are often called
GA.

\subsection{Evolution Strategies}
%----------------------------------
\label{ES}
Evolution strategies (ES) \cite{Rechenberg,Schwefel}
are typically applied to real-valued
parameter optimization problems (historically discretized). 
ES apply to real-valued vectors using Gaussian mutation, no selection
and $(\mu \stackrel{+}{,} \lambda)$ replacement strategies. 
Crossover (historically absent)  is performed either by exchanging
components, or by doing a linear recombination on some components. 

The characteristic feature of ES lies in the 
self-adaptation of the standard deviation
of the Gaussian distribution used in the mutation \cite{Baeck-Schwefel}. 
The basic idea is to add these parameters to
the genotypes, and have them  undergo evolution themselves.

\subsection{Evolutionary Programming}
%----------------------------------
\label{EP}
%It is hard to say what evolutionary programmming (EP) is because EP
%mutated during the last decade. 
Traditional EP \cite{Fogel-pere} was concerned with
evolving finite state automata for machine learning
tasks. Representation and operators were specialized for this
application area. Each parent gave birth by mutation only to one
offspring, and a plus replacement strategy was used to remove half of
the individuals.
Contemporary EP, however, \cite{DFogel} 
has evolved to using any representation and
different evolution engines, and nowadays differs from ES by using
a stochastic form of the plus replacement strategy, and by never using
crossover  (EP also uses self-adaptation of Gaussian mutation in the
case of real-valued genotypes).

\subsection{Genetic Programming}
%----------------------------------
\label{GP}
The youngest brother of the family \cite{Koza,Banzhaf:book}
has a specific application area in
machine learning and modeling tasks. A natural representation is that
of parse-trees of formal logical expressions describing a model or
procedure. Crossover and mutation operators are adapted so that they
work on trees (with varying sizes). 
Evolution engine is "inherited" from GAs (GP has long been seen as
GA with tree representation). On the other hand, syntactic expressions
-- for instance LISP --
can be viewed as programs, which makes GP the branch concerned with
automatic evolution of programs.  

\section{Application areas}
%%%%%%%%%%%%%%%%%%%%%%%%%%%%%%%%%%%%%%%%%%%%%%%%%%%%%%
\label{applications}

Although it is often stressed that an evolutionary algorithm is not an
optimizer in the strict sense \cite{DeJong:PPSN92}, optimization
problems form the most important application area of EAs. Within this
field further distinctions can be made, 
combinatorial optimization, continuous parameter optimization, or
mixed discrete-continuous optimization. 

In the framework of {\bf combinatorial optimization}, it is now recognized
that EC alone is not competitive \cite{HEC} compared to classical
Operational Research heuristics. However, hybridization of EC with those
specialized OR heuristics gave tremendous results, on benchmark
problems (e.g. best-to-date results on some difficult {\em graph coloring},
{\em quadratic assignment}, or {\em constraint satisfaction}
instances) as well as on many real-world problems 
(e.g. {\em time-tabling} in universities, {\em crew scheduling} in big
  companies, multiple tours with time-windows in {\em distribution
    applications},
\ldots). It is worth mentioning here that combinatorial problems is
today the most profitable application domain for EC.

When it comes to {\bf continuous parametric optimization}, the mistake
to avoid is to 
try to compete with highly performing numerical methods. However, in
many cases such methods do not apply (lack of regularity) or fail
(high multi-modularity). In such contexts, EC has been successfully
used for {\em control, electromagnetism, fluid mechanics, structural
analysis, \ldots }

The flexibility of EC allows one to handle representations (section
\ref{representation}) that are out of reach of any other method. This
is the case for mixed search spaces (with both discrete and
continuous variables), and even more for {\bf variable length
representations} (e.g. parse trees of Genetic Programming, see section
\ref{GP}). And this opens up the possibility for huge improvements in
areas such as {\em Machine Learning}
(e.g. by evolving sets of rules, cellular automata rules, \ldots), 
{\em modeling} (in the general framework of function identification),
{\em design} and {\em art} \cite{Bentley:book:99},
where restricting the representation of the 
solutions to a fixed set of parameters definitely bias the search
toward poo regions in terms of diversity.

Finally, let us stress that one domain where Evolutionary Algorithms
encounter an increasing attention is that of {\bf multi-objective
optimization}: specific selection  methods \cite{Deb:EMO:book} 
allow one to
spread the population of an EA over the {\em Pareto front} of a
multi-objective problem (the set of the best compromises between
the objectives), requiring only a fraction of
computing time more than the optimization of a single objective.

\section{Concluding remarks}
%%%%%%%%%%%%%%%%%%%%%%%%%%%%%
\label{conclusion}
Natural evolution can be considered as a powerful problem solver
achieving Homo Sapiens from chaos in only a couple of billion
years. Computer-based evolutionary processes can also be used as
efficient problem solvers for optimization, constraint handling,
machine learning and modeling tasks. Furthermore, many real-world
phenomena from the study of life, economy, and society can be
investigated by simulations based on evolving systems. Last but not
least, evolutionary art and design form an emerging field of
applications of  the Darwinian ideas. We expect that computer
applications based on evolutionary principles will gain popularity in
the coming years in science, business, and entertainment.  

{\small
% \bibliographystyle{plain}
% \bibliography{LA_TOTALE}
% \bibliography{../Bib/LA_TOTALE}

}
\end{document}